\pdfoutput=1

\documentclass[11pt]{article}
\usepackage{ascmac}
\usepackage{algorithmic}
\usepackage{algorithm}
\usepackage{graphicx}

\usepackage[preprint]{acl}

\usepackage{times}
\usepackage{latexsym}

\usepackage[T1]{fontenc}

\usepackage[utf8]{inputenc}

\usepackage{microtype}

\usepackage{inconsolata}

\title{70B-parameter large language models in Japanese medical question-answering}

\author{Issey Sukeda \hspace{1cm} Risa Kishikawa \hspace{1cm} Satoshi Kodera \\
        Department of Cardiovascular Medicine, Graduate School of Medicine, The University of Tokyo\\}

\begin{document}
\maketitle
\begin{abstract}
Since the rise of large language models (LLMs), the domain adaptation has been one of the hot topics in various domains.
Many medical LLMs trained with English medical dataset have made public recently. 
However, Japanese LLMs in medical domain still lack its research 
Here we utilize multiple 70B-parameter LLMs for the first time and show that instruction tuning using Japanese medical question-answering dataset significantly improves the ability of Japanese LLMs to solve Japanese medical license exams, surpassing 50\% in accuracy. 
In particular, the Japanese-centric models exhibit a more significant leap in improvement through instruction tuning compared to their English-centric counterparts. This underscores the importance of continual pretraining and the adjustment of the tokenizer in our local language.
We also examine two slightly different prompt formats, resulting in non-negligible performance improvement.
\end{abstract}

\section{Introduction}

In recent years, there has been a growing number of large language models (LLMs) specializing in a specific domain such as finance~\cite{xie2023pixiu}~\cite{Xie2023FinPythia} and medicine. In medical domain, while non-public models, such as Med-PaLM2~\cite{singhal2023large} and GPT-4 with prompting techniques~\cite{nori2023medprompt}, have achieved the state of the art in medical question-answering tasks, open-source efforts have been also made to achieve comparable results in some tasks.
For instance, PMC-LLaMA~\cite{wu2023pmcllama}, having 7B or 13B parameters, is developed by pretraining LLaMA~\cite{touvron2023llama} on 4.8M PubmedCentral papers and Medical Books.
MEDITRON-70B~\cite{chen2023meditron70b} is a continual pretrained model derived from Llama 2~\cite{touvron2023llama2} using approximately  50B tokens of medical articles, which currently holds the position of the largest medical LLM among public models.

On the other hand, the capabilities and limitations of medical LLMs in Japanese contexts remain largely unexplored. The performance of GPT-4 in the Japanese National Medical License Exam (NMLE) has been investigated, and while it already exceeds the passing standard, there have been reports of selecting forbidden choices in some questions~\cite{kasai2023evaluating}. However, except for JMedLoRA~\cite{sukeda2023jmedlora}, which is based on Llama 2 and represents the initial attempt at instruction tuning in Japanese medical articles focusing on two different domain adaptations – one in medicine and the other in language – no other research has been conducted. Our work is the first to apply multiple 70B-parameter LLMs in Japanese medical domain adaptation, resulting in the development of the currently strongest Japanese LLM particularly excelling in the domain of medical question-answering. \footnote{Our developed model is released at \url{https://huggingface.co/AIgroup-CVM-utokyohospital/MedSwallow-70b}.}.

\begin{figure}[t]
    \centering
    \includegraphics[width=1\linewidth]{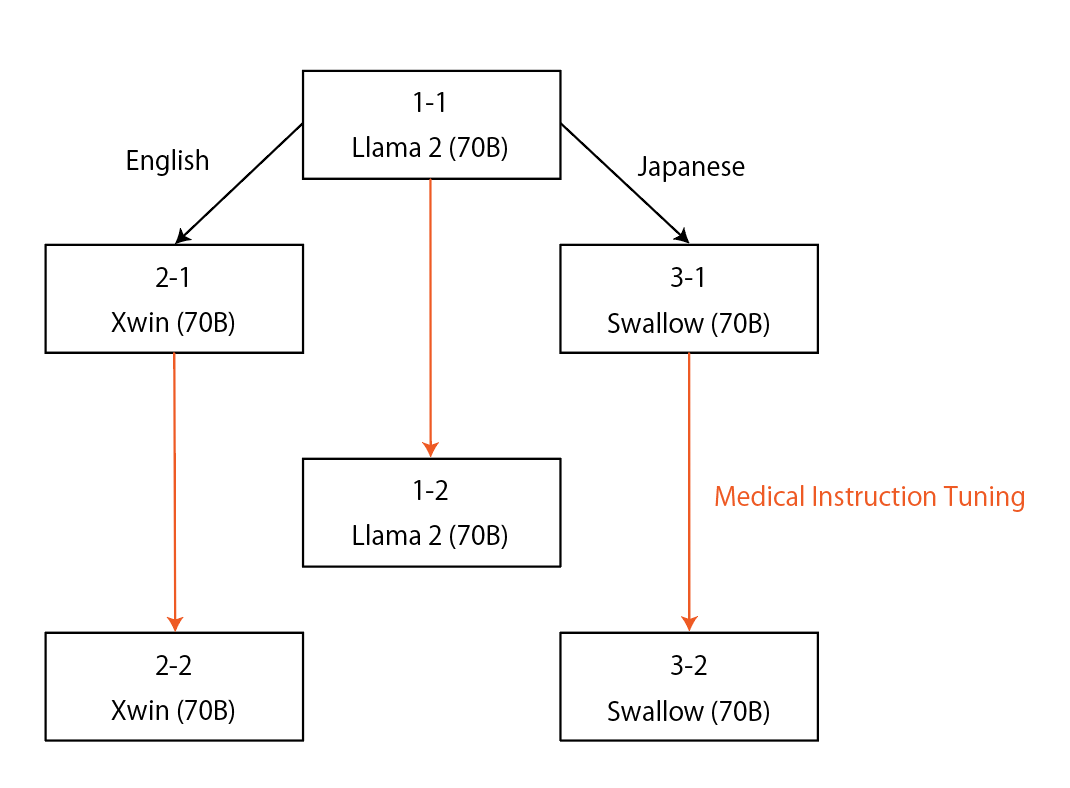}
    \caption{Overview of our candidate LLMs}
    \label{fig:overall}
\end{figure}

Our main findings are two-folds.
Firstly, while instruction tuning in a Japanese question-answer dataset consistently contributes to performance improvement in every setting, 
a Japanese continual-pretrained LLM yields better results than an English one for answering medical questions, surpassing 50\% in accuracy. 
These results are consistent with the idea that the superior performance when based on continual-pretraining in Japanese is attributed to the substantial inclusion of Japanese data in the pretraining process, and the tokenizer being optimized for Japanese processing.

Secondly, while preparing two similar prompts, there was a reasonably significant gap in accuracy, reaching up 8\% in some cases. This result indicates that even the differences between prompts that are nearly synonymous are not negligible.

\section{Medical Instruction Tuning in Japanese}
\begin{table}[t]
    \centering
    \begin{tabular}{cccc}
       \textbf{\#ID} & \textbf{Base model} & \textbf{Instruction tuning} &\textbf{} \\ \hline
       1-1& Llama 2 & none & \\ \hline
       1-2& Llama 2 & 3000 steps \\\hline
       2-1& Xwin & none & \\ \hline
       2-2& Xwin & 3000 steps \\\hline
       3-1& Swallow & none & \\ \hline
       3-2& Swallow & 3000 steps \\\hline
       4& GPT-4 & none \\\hline
    \end{tabular}
    \caption{Model settings in our experiments}
    \label{tab:models}
\end{table}

Our research is devoted to examining the performance of several 70B-parameter LLMs, which are the largest among the available models, in medical question-answering. We perform instruction tuning using medical texts on different base models, as summarized in Table~\ref{tab:models} and Figure~\ref{fig:overall}. GPT-4\footnote{\url{https://openai.com/gpt-4}} is added as \#4 for reference.

\subsection{Base Model}

All of our experiments are built on Llama 2 and its variants.
Llama 2~\cite{touvron2023llama2} with 65B parameters has been the baseline model in open-source community since its release by Meta Inc. 
In addition, we employ \textit{Xwin-LM-70B-V0.1}~\cite{xwin-lm}, which is hereafter referred to as Xwin in this paper.
Although the details of this model is not made public, Xwin is reported to outperform GPT-4~\cite{achiam2023gpt4} on AlpacaEval benchmark~\cite{alpaca_eval}.
We also use the currently most powerful Japanese LLM \textit{Swallow-70b-instruct-hf}\footnote{\url{https://huggingface.co/tokyotech-llm/Swallow-70b-instruct-hf}}
, which is hereafter referred to as Swallow in this paper. 
Both of Xwin and Swallow have undergone continual-pretraining from Llama 2 in English and Japanese resources, repspectively.

\subsection{QLoRA}

QLoRA~\cite{dettmers2023qlora} is one of the parameter efficient fine-tuning method of LLMs, incorporating quantization into low rank adaptation (LoRA)~\cite{hu2021lora}. Hyperparameters we used are listed in Appendix~\ref{appendix:qlora-parameters}.

\subsection{Instruction Dataset}
To conduct instruction tuning on each model, we prepare \textbf{USMLE-JP}, 12723 records from the United States Medical Licensing Examination(USMLE)~\cite{jin2021disease}, where all the questions, choices, and answers are translated in Japanese by Japanese medical doctors by hand.
During the medical instruction tuning phase, 
English Alpaca prompt~\cite{alpaca} is employed.



\section{Evaluation}
\subsection{Evaluation Dataset}
The questions from NMLE in 2018 is used for evaluation, which is made public online as IgakuQA~\cite{kasai2023evaluating}. The number of questions is 277 and the question format is a 5-choice structure (see Appendix~\ref{appendix:igakuqa}).

Throughout the evaluation, 1-shot Chain-of-Thought (CoT) prompting~\cite{wei2022chain} is applied for inference in two slightly different ways : one follows Med-PaLM2~\cite{singhal2023towards} and another follows Alpaca~\cite{alpaca}.
These two prompts only differ in the order of sentences
(see Appendix~\ref{appendix:prompt-format}).





\subsection{Metrics}

Sukeda et al.~\cite{sukeda2023jmedlora} uses three different metrics: Exact match, Gestalt score, and Accuracy. These metrics calculate the discrepancy between the correct choice and the model's output. While Exact match does not allow any slight misspecification in any tokens,  Gestalt score and Accuracy are based on Gestalt distance calculated by pattern matching algorithm and robust to such issues.
However, this approach has two weakness: (i) it is prone to the slight misspecification of each token in the output (ii) it does not evaluate with regard to the order for questions that involve selecting multiple choices.

Here we have made a slight update in the definition of Accuracy and adopted it as our evaluation metric. Algorithm~\ref{alg1} shows the procedure of calculating is\_correct for each question. Accuracy is defined as the average of is\_correct.

\begin{figure}[!t]
\begin{algorithm}[H]
    \caption{Evaluation of the correctness for each question-answer pair}
    \label{alg1}
    \begin{algorithmic}[]
    \REQUIRE $\mathcal{C}$ : choices, $C^*$ : correct choices, $R$ : model's output, $G(\cdot,\cdot)$ : Gestalt distance
    \IF{$|C^*| = 1$}
    \STATE is\_correct = 1 if $C^* = \mathrm{argmax}_{C \in \mathcal{C}} G(C,R)$ else 0
    \ELSE[$|C^*| = 2$]
    \STATE $R_1, R_2 \leftarrow \mathrm{split}(R)$
    \STATE $C_1 \leftarrow \mathrm{argmax}_{C \in \mathcal{C}} G(C,R_1)$
    \STATE $C_2 \leftarrow \mathrm{argmax}_{C \in \mathcal{C}} G(C,R_2)$
    \STATE is\_correct = 1 if $C^* = \{C_1,C_2\}$ else 0
    \ENDIF
    \RETURN is\_correct
    \end{algorithmic}
\end{algorithm}
\end{figure}


\section{Results}
\begin{table*}[!t]
    \centering
    \begin{tabular}{ccccccc} 
       \textbf{\#Model ID} &\textbf{Prompt}&\textbf{Correct} &\textbf{Invalid} & \textbf{Wrong} &\textbf{Accuracy}&\textbf{Improvement} \\ \hline
       1-1  &(A)&53&9&215&0.191&-\\\hline
       1-1  &(B)&45&7&225&0.162&-\\\hline
       1-2  &(A)& 89&14&174&0.321&+ 0.130\\\hline
       1-2  &(B)& 94&28&155&0.339&+ 0.177\\\hline
       2-1  &(A)& 102&2 & 173&0.368&(\#1-1) + 0.177\\ \hline
       2-1  &(B)& 87&8 & 182&0.314&(\#1-1) + 0.152\\ \hline
       2-2  &(A)& 103&27& 147&0.372&+ 0.004\\\hline
       2-2  &(B)& 117&25& 135&0.422&+ 0.108\\\hline
       3-1  &(A)& 50&14&213&0.180&(\#1-1) $-$ 0.010\\ \hline
       3-1  &(B)& 74&5&198&0.267&(\#1-1) + 0.105\\ \hline
       3-2  &(A)& 144& 10& 123& \textbf{0.519}&+ 0.339\\\hline
       3-2  &(B)& 134& 11& 132 & \textbf{0.484}&+ 0.217\\\hline
       4$^*$&(A)&31&0&6&\textbf{0.838}&-\\\hline
    \end{tabular}\\
    $^*$ The number of evaluation dataset is reduced due to computational cost.
    \caption{Performance results. Xwin and Swallow are compared with Llama 2 to quantify the contribution of continual pretraining. Each of the models after QLoRA is compared with its base model.}
    \label{tab:results-full}
\end{table*}


\begin{figure}[t]
    \centering
    \includegraphics[width=\linewidth]{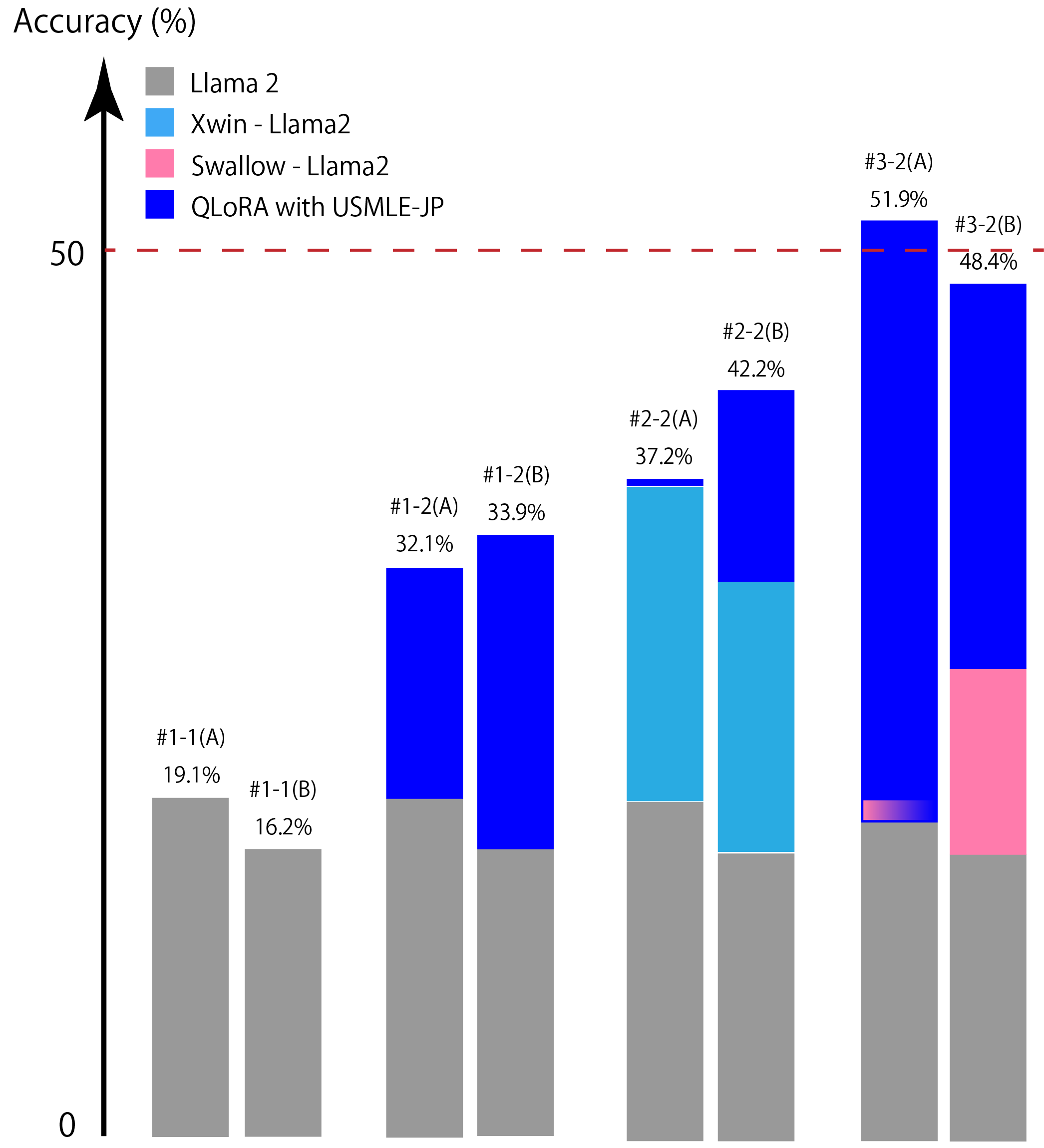}
    \caption{Improvement by QLoRA instruction tuning in Accuracy. Gray shows the performance of Llama 2 as baseline. Light blue shows the difference between Xwin (original) and Llama 2 (original). Pink shows the difference between Swallow (original) and Llama 2 (original), which is negative in \#3-2(A). Blue shows the contribution of QLoRA.}
    \label{fig:score-bar}
\end{figure}

Table~\ref{tab:results-full} shows the performance of each model in answering IgakuQA 2018 by single run. 
Incorrect responses include \textbf{Invalid} responses, where the number in instruction and the number of choices in model's output are not equal, and \textbf{Wrong} responses, where the model simply choose wrong answer. Top-3 Accuracy is emphasized in bold.
In the \textbf{Improvement} column, the original Xwin and Swallow are compared with Llama 2 to quantify the contribution of continual pretraining. Each of the other models is compared with its base model to quantify the contribution of QLoRA.

\subsection{Base Model Selection : Swallow outperforms Xwin}
First we argue that the base model more suited to the target task is more preferable.
When comparing the best performances of each model, Swallow performed better than Xwin, followed by Llama 2, around 9\% difference each. This result exhibits the effect of suited continual pretraining. 
Two indistinguishable and mutually related factors are the base model improvement and the tokenizer improvement. 
Evidently, Swallow passes continual pretraining with more than 90B tokens~\cite{fuji-nakamura-etal-2024-swallow-llm}, thus its ability in Japanese should be better than English-centric Xwin. In addition, since Swallow is intended to solve Japanese tasks, its tokenizer is optimized mainly for Japanese. Figure~\ref{fig:score-bar} illustrates that while the enhancement by QLoRA on Swallow is substantial, the original Swallow is not quite competitive --- even worse than Llama 2 when prompt (A) is used. This trend is in contrast with the results for Xwin, suggesting that the improvement and adjustment in its tokenizer contributes more to the performance increase than the improvement in the base model.

Moreover, it is observed that Llama 2 and Xwin output more invalid responses after instruction tuning compared to Swallow. Most of these invalid responses included only one choice as the answer, implying a deterioration in the ability to capture numbers mentioned in instructions properly when English-centric models are finetuned in Japanese.

\subsection{Format of CoT Prompts}

Should the CoT prompt follow Med-PaLM2~\cite{singhal2023towards} or Alpaca~\cite{alpaca}? These two prompts have almost the same meaning but differ slightly in how they instruct the model. Table~\ref{tab:results-full} demonstrates that this difference resulted in a non-negligible accuracy gap as large as 8.7\% at most.

In our experiments \#1-1, \#2-1, and \#3-2, prompt (A) outperforms prompt (B) in accuracy, while the opposite is true in the rest of the cases.
Which prompt is preferable depends on the situation, regardless of the type of base model or the presence of tuning.
This observation, indicating that accuracy varies due to slight differences in prompts, highlights the difficulty of establishing a unified approach to constructing domain-specific LLMs.

\subsection{Comparison with GPT-4}

In our experimental settings, neither Xwin nor Swallow achieved the level of accuracy exhibited by the original GPT-4, with an approximate 30\% gap, even after instruction tuning specific to the medical domain. As in Table~\ref{tab:swallow-vs-gpt}, there was only one question where our best model, namely \#3-2,  provided a correct answer while GPT-4 made an incorrect response.
Remarkably, GPT-4 did not generate invalid response at all.

\begin{table}
    \centering
    \begin{tabular}{c|cc}
         & \begin{tabular}{c}
             Correct\\(Swallow) 
         \end{tabular} & \begin{tabular}{c}
         Wrong\\(Swallow) \end{tabular}\\ \hline
         Correct(GPT-4)&12  &19 \\
         Wrong(GPT-4)& 1 & 5\\
    \end{tabular}
    \caption{Swallow(\#3-2, (A)) vs GPT(\#4, (A)) in a subset of IgakuQA 2018.}
    \label{tab:swallow-vs-gpt}
\end{table}

\subsection{Limitations and Future Works}


Using multiple-choice questions in the evaluation of LLM has been controversial~\cite{pezeshkpour2023large}~\cite{zheng2023large}. In Appendix~\ref{appendix:usmlejp-eval}, we demonstrate the fact that the score significantly drops after the shuffle of choices. Further exploration is required to determine the most meaningful evaluation metrics.

The size of the training and evaluation datasets is limited. Our work suggests significant benefits of training in the local language, emphasizing the importance of curating the available Japanese medical corpus to construct a practical and useful LLM in a local environment such as clinics.

Also, the validity of training with USMLE and evaluating on NMLE should be further argued sicne both of them are medical license exams but in different countries and languages.

Furthermore, it has been noted that prompt engineering significantly impacts the performance of LLMs, although this was beyond the scope of our research. Utilizing multiple-shot inference, self-consistency~\cite{wang2022self}, ensemble refinement~\cite{singhal2023towards}, and Medprompt~\cite{nori2023medprompt} may lead to a significant improvement in their performance also in Japanese context.

\section{Conclusion}

Our work has demonstrated the possibility and limitations of the best accessible model that we can construct locally in each clinical institution, focusing on medical domain adaptation and Japanese adaptation simultaneously. 
Compared to its English-centric counterparts, the use of the currently strongest Japanese LLM as base model has amplified the effect of instruction tuning. When using Med-PaLM2-like CoT prompting, the performance in Japanese medical question-answering has substantially increased, surpassing 50\% in accuracy. 


\section*{Acknowledgements}

This study was supported by the Japan Agency for Medical Research and Development (Grant Number: JP23hk0102078h0003).
We thank Dr. Hisahiko Sato for creating and providing us with USMLE-JP dataset.

\section*{Ethical Consideration}

We intend not to use our models for any clinical purposes, but only for research purposes.

\bibliography{custom}

\begin{thebibliography}{23}
\expandafter\ifx\csname natexlab\endcsname\relax\def\natexlab#1{#1}\fi

\bibitem[{Chen et~al.(2023)Chen, Hernández-Cano, Romanou, Bonnet, Matoba, Salvi, Pagliardini, Fan, Köpf, Mohtashami, Sallinen, Sakhaeirad, Swamy, Krawczuk, Bayazit, Marmet, Montariol, Hartley, Jaggi, and Bosselut}]{chen2023meditron70b}
Zeming Chen, Alejandro Hernández-Cano, Angelika Romanou, Antoine Bonnet, Kyle Matoba, Francesco Salvi, Matteo Pagliardini, Simin Fan, Andreas Köpf, Amirkeivan Mohtashami, Alexandre Sallinen, Alireza Sakhaeirad, Vinitra Swamy, Igor Krawczuk, Deniz Bayazit, Axel Marmet, Syrielle Montariol, Mary-Anne Hartley, Martin Jaggi, and Antoine Bosselut. 2023.
\newblock \href {http://arxiv.org/abs/2311.16079} {Meditron-70b: Scaling medical pretraining for large language models}.

\bibitem[{Dettmers et~al.(2023)Dettmers, Pagnoni, Holtzman, and Zettlemoyer}]{dettmers2023qlora}
Tim Dettmers, Artidoro Pagnoni, Ari Holtzman, and Luke Zettlemoyer. 2023.
\newblock Qlora: Efficient finetuning of quantized llms.
\newblock \emph{arXiv e-prints}, pages arXiv--2305.

\bibitem[{Fujii et~al.(2024)Fujii, Nakamura, Mengsay, Iida, Oi, Hattori, Hirai, Mizuki, Yokota, and Okazaki}]{fuji-nakamura-etal-2024-swallow-llm}
Kazuki Fujii, Taishi Nakamura, Loem Mengsay, Daiki Iida, Seiya Oi, Sho Hattori, Shota Hirai, Sae Mizuki, Rio Yokota, and Naohiro Okazaki. 2024.
\newblock Building a robust large-language model in japanese through continual pretraining : keizokujizengakushu ni yoru nihongo ni tuyoi daikibogengomoderu no koutiku, in {J}apanese.
\newblock In \emph{NLP2024}.

\bibitem[{Hu et~al.(2021)Hu, Wallis, Allen-Zhu, Li, Wang, Wang, Chen et~al.}]{hu2021lora}
Edward~J Hu, Phillip Wallis, Zeyuan Allen-Zhu, Yuanzhi Li, Shean Wang, Lu~Wang, Weizhu Chen, et~al. 2021.
\newblock Lora: Low-rank adaptation of large language models.
\newblock In \emph{International Conference on Learning Representations}.

\bibitem[{Jin et~al.(2021)Jin, Pan, Oufattole, Weng, Fang, and Szolovits}]{jin2021disease}
Di~Jin, Eileen Pan, Nassim Oufattole, Wei-Hung Weng, Hanyi Fang, and Peter Szolovits. 2021.
\newblock What disease does this patient have? a large-scale open domain question answering dataset from medical exams.
\newblock \emph{Applied Sciences}, 11(14):6421.

\bibitem[{Kasai et~al.(2023)Kasai, Kasai, Sakaguchi, Yamada, and Radev}]{kasai2023evaluating}
Jungo Kasai, Yuhei Kasai, Keisuke Sakaguchi, Yutaro Yamada, and Dragomir Radev. 2023.
\newblock Evaluating {GPT}-4 and {ChatGPT} on japanese medical licensing examinations.
\newblock \emph{arXiv preprint arXiv:2303.18027}.

\bibitem[{Li et~al.(2023)Li, Zhang, Dubois, Taori, Gulrajani, Guestrin, Liang, and Hashimoto}]{alpaca_eval}
Xuechen Li, Tianyi Zhang, Yann Dubois, Rohan Taori, Ishaan Gulrajani, Carlos Guestrin, Percy Liang, and Tatsunori~B. Hashimoto. 2023.
\newblock Alpacaeval: An automatic evaluator of instruction-following models.
\newblock \url{https://github.com/tatsu-lab/alpaca_eval}.

\bibitem[{Nori et~al.(2023)Nori, Lee, Zhang, Carignan, Edgar, Fusi, King, Larson, Li, Liu et~al.}]{nori2023medprompt}
Harsha Nori, Yin~Tat Lee, Sheng Zhang, Dean Carignan, Richard Edgar, Nicolo Fusi, Nicholas King, Jonathan Larson, Yuanzhi Li, Weishung Liu, et~al. 2023.
\newblock Can generalist foundation models outcompete special-purpose tuning? case study in medicine.
\newblock \emph{arXiv preprint arXiv:2311.16452}.

\bibitem[{OpenAI(2023)}]{achiam2023gpt4}
OpenAI. 2023.
\newblock {GPT-4 Technical Report}.
\newblock \emph{arXiv preprint arXiv:303.08774}.

\bibitem[{Pezeshkpour and Hruschka(2023)}]{pezeshkpour2023large}
Pouya Pezeshkpour and Estevam Hruschka. 2023.
\newblock {Large language models sensitivity to the order of options in multiple-choice questions}.
\newblock \emph{arXiv preprint arXiv:2308.11483}.

\bibitem[{Singhal et~al.(2023{\natexlab{a}})Singhal, Azizi, Tu, Mahdavi, Wei, Chung, Scales, Tanwani, Cole-Lewis, Pfohl et~al.}]{singhal2023large}
Karan Singhal, Shekoofeh Azizi, Tao Tu, S~Sara Mahdavi, Jason Wei, Hyung~Won Chung, Nathan Scales, Ajay Tanwani, Heather Cole-Lewis, Stephen Pfohl, et~al. 2023{\natexlab{a}}.
\newblock Large language models encode clinical knowledge.
\newblock \emph{Nature}, pages 1--9.

\bibitem[{Singhal et~al.(2023{\natexlab{b}})Singhal, Tu, Gottweis, Sayres, Wulczyn, Hou, Clark, Pfohl, Cole-Lewis, Neal et~al.}]{singhal2023towards}
Karan Singhal, Tao Tu, Juraj Gottweis, Rory Sayres, Ellery Wulczyn, Le~Hou, Kevin Clark, Stephen Pfohl, Heather Cole-Lewis, Darlene Neal, et~al. 2023{\natexlab{b}}.
\newblock Towards expert-level medical question answering with large language models.
\newblock \emph{arXiv preprint arXiv:2305.09617}.

\bibitem[{Sukeda et~al.(2023)Sukeda, Suzuki, Sakaji, and Kodera}]{sukeda2023jmedlora}
Issey Sukeda, Masahiro Suzuki, Hiroki Sakaji, and Satoshi Kodera. 2023.
\newblock {JMedLoRA: Medical Domain Adaptation on Japanese Large Language Models using Instruction-tuning}.
\newblock \emph{arXiv preprint arXiv:2310.10083}.

\bibitem[{Taori et~al.(2023)Taori, Gulrajani, Zhang, Dubois, Li, Guestrin, Liang, and Hashimoto}]{alpaca}
Rohan Taori, Ishaan Gulrajani, Tianyi Zhang, Yann Dubois, Xuechen Li, Carlos Guestrin, Percy Liang, and Tatsunori~B. Hashimoto. 2023.
\newblock {Stanford Alpaca: An Instruction-following LLaMA model}.
\newblock \url{https://github.com/tatsu-lab/stanford_alpaca}.

\bibitem[{Touvron et~al.(2023{\natexlab{a}})Touvron, Lavril, Izacard, Martinet, Lachaux, Lacroix, Rozi{\`e}re, Goyal, Hambro, Azhar et~al.}]{touvron2023llama}
Hugo Touvron, Thibaut Lavril, Gautier Izacard, Xavier Martinet, Marie-Anne Lachaux, Timoth{\'e}e Lacroix, Baptiste Rozi{\`e}re, Naman Goyal, Eric Hambro, Faisal Azhar, et~al. 2023{\natexlab{a}}.
\newblock {Llama: Open and efficient foundation language models}.
\newblock \emph{arXiv preprint arXiv:2302.13971}.

\bibitem[{Touvron et~al.(2023{\natexlab{b}})Touvron, Martin, Stone, Albert, Almahairi, Babaei, Bashlykov, Batra, Bhargava, Bhosale et~al.}]{touvron2023llama2}
Hugo Touvron, Louis Martin, Kevin Stone, Peter Albert, Amjad Almahairi, Yasmine Babaei, Nikolay Bashlykov, Soumya Batra, Prajjwal Bhargava, Shruti Bhosale, et~al. 2023{\natexlab{b}}.
\newblock {Llama 2: Open foundation and fine-tuned chat models}.
\newblock \emph{arXiv preprint arXiv:2307.09288}.

\bibitem[{Wang et~al.(2022)Wang, Wei, Schuurmans, Le, Chi, Narang, Chowdhery, and Zhou}]{wang2022self}
Xuezhi Wang, Jason Wei, Dale Schuurmans, Quoc Le, Ed~Chi, Sharan Narang, Aakanksha Chowdhery, and Denny Zhou. 2022.
\newblock Self-consistency improves chain of thought reasoning in language models.
\newblock \emph{arXiv preprint arXiv:2203.11171}.

\bibitem[{Wei et~al.(2022)Wei, Wang, Schuurmans, Bosma, Xia, Chi, Le, Zhou et~al.}]{wei2022chain}
Jason Wei, Xuezhi Wang, Dale Schuurmans, Maarten Bosma, Fei Xia, Ed~Chi, Quoc~V Le, Denny Zhou, et~al. 2022.
\newblock Chain-of-thought prompting elicits reasoning in large language models.
\newblock \emph{Advances in Neural Information Processing Systems}, 35:24824--24837.

\bibitem[{Wu et~al.(2023)Wu, Lin, Zhang, Zhang, Wang, and Xie}]{wu2023pmcllama}
Chaoyi Wu, Weixiong Lin, Xiaoman Zhang, Ya~Zhang, Yanfeng Wang, and Weidi Xie. 2023.
\newblock {PMC-LLaMA: Towards Building Open-source Language Models for Medicine}.
\newblock \emph{arXiv preprint arXiv:2304.14454}.

\bibitem[{Xie et~al.(2023)Xie, Han, Zhang, Lai, Peng, Lopez-Lira, and Huang}]{xie2023pixiu}
Qianqian Xie, Weiguang Han, Xiao Zhang, Yanzhao Lai, Min Peng, Alejandro Lopez-Lira, and Jimin Huang. 2023.
\newblock Pixiu: A large language model, instruction data and evaluation benchmark for finance.
\newblock \emph{arXiv preprint arXiv:2306.05443}.

\bibitem[{{Xwin-LM\ Team}(2023)}]{xwin-lm}
{Xwin-LM\ Team}. 2023.
\newblock \href {https://github.com/Xwin-LM/Xwin-LM} {{Xwin-LM}}.

\bibitem[{Yong et~al.(2023)Yong, Karana, and Aitzaz}]{Xie2023FinPythia}
Xie Yong, Aggarwal Karana, and Ahmad Aitzaz. 2023.
\newblock Efficient continual pre-training for building domain specific large language models.
\newblock \emph{https://arxiv.org/pdf/2311.08545.pdf}.

\bibitem[{Zheng et~al.(2023)Zheng, Zhou, Meng, Zhou, and Huang}]{zheng2023large}
Chujie Zheng, Hao Zhou, Fandong Meng, Jie Zhou, and Minlie Huang. 2023.
\newblock {Large Language Models Are Not Robust Multiple Choice Selectors}.
\newblock \emph{arXiv preprint arXiv:2309.03882}.

\end{thebibliography}

\appendix

\section{QLoRA Hyperparameters}\label{appendix:qlora-parameters}
QLoRA~\cite{dettmers2023qlora} is one of the parameter efficient fine-tuning method of LLMs, incorporating quantization into low rank adaptation (LoRA)~\cite{hu2021lora}. Hyperparameters we used throughout our experiments are listed in Table~\ref{tab:qlora-parameters}.

\begin{table}[h]
  \centering
  \caption{Hyperparameters for QLoRA}
  \scalebox{1.0}{
  \begin{tabular}{|c|c|}\hline
    learning rate &2e-4\\
    input length&512\\
    target max length&512\\
    batch size&16\\
    max steps& 3000\\
    $r$ of QLoRA&64\\
    $\alpha$ of QLoRA&16\\
    dropout rate of QLoRA&0.1\\
    target parameter&all linear layers\\
    \hline
  \end{tabular}
  }
  \label{tab:qlora-parameters}
\end{table}

\section{Details of IgakuQA dataset}\label{appendix:igakuqa}

IgakuQA~\cite{kasai2023evaluating} includes Japanese Medical License Exams from 2018 to 2022. 
The 2018 exam includes a total of 400 five-choices questions. In this study, as LLMs can only handle text, we decided to use a subset consisting of 284 text-only questions. However, there were 7 questions that required selecting three or more options, and due to their complexity, we excluded them. As a result, we utilized the remaining 277 questions for experiments.

\section{Prompt Formats}\label{appendix:prompt-format}

Two slightly different prompt formats in 1-shot manner are applied in evaluation to observe its influence on performances. 
Prompt (A) follows Med-PaLM2~\cite{singhal2023towards}, the best medical LLM. Prompt (B) follows Alpaca~\cite{alpaca}, aligning with the instruction tuning step. 
For both prompt formats, questions are input in \{instruction\} and choices are input in \{input\}.

\begin{itembox}[l]{CoT prompt (A) (originally in Japanese)}
\#\#\# Instruction:\\
The following are multiple choice questions about medical knowledge. Solve them in a step-by-step fashion,
starting by summarizing the available information. Output a single option from the five options as the final answer.\\
\#\#\# Input:\\
\{instruction\}\\
\{input\}\\
\#\#\# Response: 
\end{itembox}

\begin{itembox}[l]{CoT prompt (B) (originally in Japanese)}
The following are multiple choice questions about medical knowledge. Solve them in a step-by-step fashion,
starting by summarizing the available information. Output a single option from the five options as the final answer.
\#\#\# Instruction:\\
\{instruction\}\\
\#\#\# Input:\\
\{input\}\\
\#\#\# Response: 
\end{itembox}

\section{Ablation Studies}

\subsection{Changing evaluation dataset into USMLE-JP} \label{appendix:usmlejp-eval}

This part is devoted to confirm that LLMs can memorize the answers contained in instruction dataset. Here, we use USMLE-JP instead of IgakuQA in 2018 for evaluation, letting the data leakage occur on purpose.

As a result,  Xwin with 3000 steps of QLoRA (\#1-3) achieved Accuracy = 0.827 using CoT prompt (A), and Accuracy = 0.822 using CoT prompt (B), respectively. We conclude that instruction tuning based on QLoRA is capable of memorising training dataset sufficiently, although not completely.

\subsection{Changing instruction dataset into medical journal articles}

We performed instruction tuning on Llama 2, Xwin, and Swallow with Japanese medical journal articles used by~\cite{sukeda2023jmedlora}. Except the dataset used, the experimental setup followed Section 2 and Section 3. 

The performances of each model are summarized in Table~\ref{tab:results-2}. 
Through these experiments, we observe an overall decrease in accuracy compared to the instruction tuning using USMLE-JP which is presented in Table~\ref{tab:results-full}, suggesting that USMLE-JP includes knowledge that is common between Japanese medical license exams and the English one to a certain extent.

\begin{table}[!t]
    \centering
    \scalebox{0.8}{
    \begin{tabular}{ccccc} 
       \textbf{\textbf{Base Model}} &\textbf{Prompt}&\textbf{Correct} &\textbf{Invalid}&\textbf{Accuracy}\\ \hline
       Llama 2&(A)&65&9&0.234\\\hline
       Llama 2&(B)&75&12&0.270\\\hline
       Xwin&(A)&91&7&0.328\\\hline
       Xwin&(B)&80&20&0.288\\\hline
       Swallow&(A)&104&2&0.375\\\hline
       Swallow&(B)&96&9&0.346\\\hline
    \end{tabular}
    }
    \caption{Performance of models finetuned with medical journal article dataset}
    \label{tab:results-2}
\end{table}

\section{Other Information}

\subsection{Model License}
All models utilized in our experiments are covered by the LLAMA 2 COMMUNITY LICENSE AGREEMENT\footnote{\url{https://github.com/facebookresearch/llama/blob/main/LICENSE}}, which are available for research use.
Since our developed model is also built upon Llama 2, it is released under the same license.

\subsection{Computational Environment}

All instruction tuning experiments are conducted on 4 NVIDIA A100 GPUs with 80GB VRAM each. 
All evalutations are conducted on 1 NVIDIA A100 GPU with 80GB VRAM.
All source codes are developed using Python and Docker on Ubuntu 20.04.
\end{document}